\documentclass[conference]{IEEEtran}
\IEEEoverridecommandlockouts

\usepackage[T1]{fontenc}
\usepackage{graphicx}
\usepackage{amsmath}
\usepackage{mathrsfs}
\usepackage{multirow}
\usepackage{enumitem}
\usepackage{caption}
\usepackage{subcaption}
\usepackage{amsfonts}
\usepackage{array}

\usepackage{color,soul,hyperref}

\begin{document}

\title{Learning Bipedal Walking on a Quadruped Robot via Adversarial Motion Priors}

\author{Tianhu Peng\textsuperscript{1}, Lingfan Bao\textsuperscript{1}, Joseph Humphreys\textsuperscript{1}, Andromachi Maria Delfaki\textsuperscript{2}, \\
Dimitrios Kanoulas\textsuperscript{2} and Chengxu Zhou\textsuperscript{2*}

\thanks{This work was supported by the Royal Society [grant number RG\textbackslash R2\textbackslash232409] and the UKRI Future Leaders Fellowship [grant number MR/V025333/1]. Please refer to the video for an overview of our framework and results at \url{https://youtu.be/JYD1RlrQRWM}.}
\thanks{For the purpose of open access, the authors have applied a Creative Commons Attribution (CC BY) licence to any Author Accepted Manuscript version arising from this submission.}
\thanks{\textsuperscript{1}School of Mechanical Engineering, University of Leeds, UK.}
\thanks{\textsuperscript{2}Department of Computer Science, University College London, UK.}%
\thanks{\textsuperscript{*}Corresponding author, {\tt\small chengxu.zhou@ucl.ac.uk}}%
}

\maketitle

\begin{abstract}
Previous studies have successfully demonstrated agile and robust locomotion in challenging terrains for quadrupedal robots. However, the bipedal locomotion mode for quadruped robots remains unverified. This paper explores the adaptation of a learning framework originally designed for quadrupedal robots to operate blind locomotion in biped mode. We leverage a framework that incorporates Adversarial Motion Priors with a teacher-student policy to enable imitation of a reference trajectory and navigation on tough terrain. Our work involves transferring and evaluating a similar learning framework on a quadruped robot in biped mode, aiming to achieve stable walking on both flat and complicated terrains. Our simulation results demonstrate that the trained policy enables the quadruped robot to navigate both flat and challenging terrains, including stairs and uneven surfaces.
\end{abstract}


\section{Introduction}
 
Legged robots exhibit superior terrain adaptability compared to their wheeled and tracked counterparts. Although quadrupedal robots are known for their stability and agility, bipedal robots offer greater flexibility by freeing the upper body for complex tasks. This flexibility suggests the potential for quadrupedal robots to walk in a bipedal gait, using the rear legs for walking and the front legs for manipulation.

The primary challenge in adapting quadruped robots for bipedal locomotion stems from their mechanical design constraints. First, unlike typical bipedal robots that have firm, flat feet, quadruped robots often feature soft, point-contact feet that inherently lack stability. Second, the rear legs of quadruped robots are not specifically designed for bipedal walking, their limited range of motion and underactuation contribute to unnatural and unstable bipedal gaits. This design mismatch explains why quadruped robots struggle with bipedal walking modes. This leads to high requirements for locomotion controllers during bipedal modes.

To achieve bipedal walking for quadruped robots, there are primarily two approaches: the model-based method and the learning-based method \cite{2024_Bao_Review_bipedal_DRL}. Model-based methods are based on highly accurate mathematical models, which have proven to be effective in executing highly dynamic motions in both quadruped and bipedal robots. However, these methods lack robustness and generalization in unseen scenarios, largely due to the difficulty of accurately modeling ground interactions and contact dynamics. In contrast learning-based methods, reinforcement learning (RL), provides a more adaptable solution by enabling the exploration of the robots' full dynamics and interactions with the environment, thus offering greater flexibility in controlling complex locomotive behaviors. Early research in RL on legged robots primarily utilized unrealistic models within physical simulators\cite{2016_GAIL_imitation_learning,2017_Xuebinpeng_animation_deeploco_DRL_hiarachysystem_referencemotion_deepconvolutionalneuralnewtwork_jointangle_soccer.,2020_zhaoming_drl_steppingstones_PPOwithactorcritic_referencefree_simulation}. In transitioning to a practical bipedal robot and learning natural and robust gaits, previous studies have primarily designed a reference-free learning framework by designing periodic composition reward \cite{2021_siekmann_sim2real_nonreference_perodicreward_DRL_e2e_LSTM_PPO_cassie} or mimicking predefined references \cite{2020_Xie_firstsim2real_,2021_UCB_hybridrobotics_sim2real_referencebased_HZD_gaitlibrary_e2epolicy_drl_Cassie_lowpassfilter,2024_Zhang_wholebody_adversarial_motion_priors}. 

Reference-free methods explore various gait patterns efficiently, while reference-based methods leverage prior knowledge to accelerate learning, resulting in efficient policy exploration and robust locomotion skills. These methods incorporate expert information and predefined reference trajectories from motion capture data or trajectory optimization (TO). Generative Adversarial Imitation Learning \cite{2016_GAIL_imitation_learning} and Adversarial motion priors (AMP) \cite{2024_Zhang_wholebody_adversarial_motion_priors} predict state transitions and evaluate the similarity between reference and agent data, promoting stable gait maintenance. AMP was implemented with a study of human reference behavior in biped robots \cite{wu2023learning,escontrela2022adversarial}, combining it with periodic rewards to promote stable gait maintenance.

\begin{figure*}
    \centering
    \includegraphics[width=0.95\linewidth]{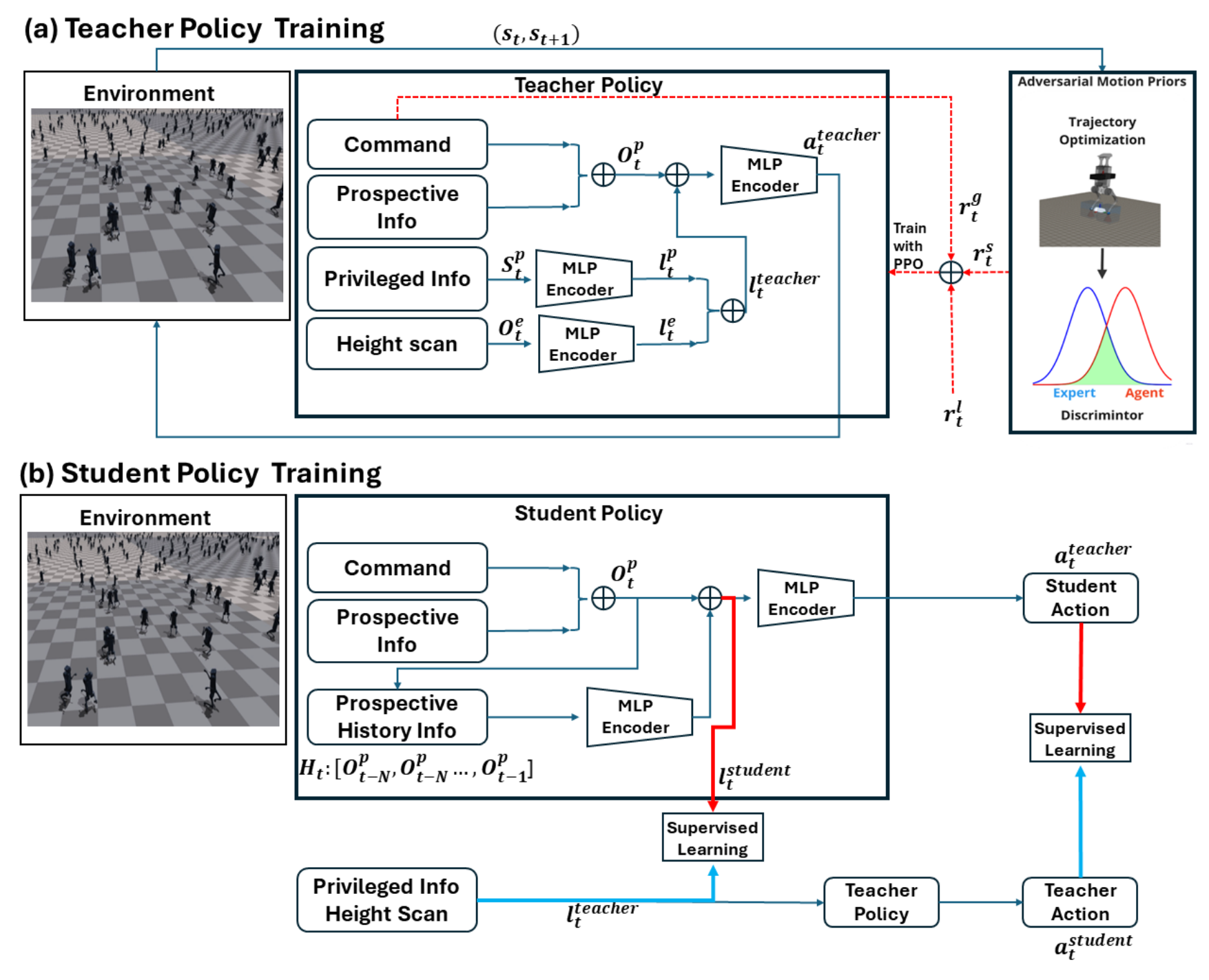}
    \caption{Overview of the teacher-student learning framework. (a) The teacher policy, which leverages privileged data \(S^p_t\),terrain information \(o^e_t\) and prospective data \(o^p_t\) through RL, aims to maximize a total reward \(r_t\) comprising command task reward \(r_t^g\), style reward \(r_t^g\) based on AMP, and regulation reward \(r_t^g\) for ensuring safety and smooth motion. (b) The student policy, trained via supervised learning, seeks to imitate the teacher's actions \(a^{teacher}_t\) and reconstruct the teacher's latent states \(l^{teacher}_t\) from historical and prospective observations \(H_t: [O_{t-N},O_{t-N-1},...,O_{t-1}]\)}
    \label{fig:Framework}
    \centering
\end{figure*}

To generate appropriate predefined references, several methods are utilized. Motion capture technology is commonly employed to produce reference data for various types of legged robots \cite{escontrela2022adversarial,wang2023learning,zhang2024whole}. This technology captures comprehensive kinematic data from real-world scenarios, enabling versatile data collection that is not confined to a specific robot model. However, adapting this data to different robotic platforms often requires a re-targeting process, which increases both complexity and manual labor. On the other hand, TO in reduced-order \cite{winkler18} and full dynamics models \cite{2021_UCB_hybridrobotics_sim2real_referencebased_HZD_gaitlibrary_e2epolicy_drl_Cassie_lowpassfilter}, has been employed. This approach reduces complexity and eliminates the need for further re-targeting, making it a more efficient solution for generating references. Additionally, compared to full dynamics models, reduced-order models can decrease the computational resources required for optimization and offer greater generalization across all quadruped robots. Regarding predesigned references from the optimization method, various models are utilized. Besides HZD-based full dynamics reference \cite{2021_UCB_hybridrobotics_sim2real_referencebased_HZD_gaitlibrary_e2epolicy_drl_Cassie_lowpassfilter}, reduced-order dynamics such as Single Rigid Body Dynamics (SRBD) \cite{2022_yu_DRL_maneuvers_SRBD_reference_based} are also utilized in the training procedure. Based on the reduced-order model, task-space learning focuses on the foot setpoints and based velocity \cite{2021_duan_DRL_task-spaceaction_hiarachycotnrolscheme_inversedynamiccontroller,2023_template_taskspace_hierarchyscheme_reducedorderstateALIP_learnedhigherlevel_lowlevelinversedynamiccontroller}. 

Another significant challenge exists in bridging the sim-to-real gap. To extend the robustness of locomotion and overcome the sim-to-real gap, frameworks using the privileged learning paradigm have been introduced \cite{lee2020learning,kumar2021rma}. By combining the strengths of AMP in reference-based learning and privileged learning, there is potential to enable quadrupedal robots to adopt bipedal walking modes. Similar frameworks have been introduced \cite{wang2023learning,wu2023learning}, but none have been validated on bipedal robots or quadruped robots for bipedal mode.

Our objective is to train a policy that enables a quadrupedal robot to achieve bipedal locomotion using only its two rear legs, thus freeing its front legs for more complex tasks. This capability aims to enhance the robot's versatility and functionality in various practical applications. This paper presents a novel framework that enables quadrupedal robots to achieve robust and agile blind bipedal locomotion on flat terrain. We adopt a teacher-student policy framework, where privileged information that the robot cannot directly access is encoded. The student policy uses historical observation information to infer this privileged information, thereby enhancing robustness. Additionally, we integrate the AMP training framework to learn and imitate the style behaviors of reference data generated through TO based on a SRBD model. Different from previous work \cite{2022_multimode_quadrupedtobipedal} with assistant devices, this comprehensive training framework equips the policy to support agile bipedal motions in quadrupedal robots.

In summary, the primary contribution of this paper is to develop a novel framework (shown in Fig. \ref{fig:Framework}) that allows quadrupedal robots to perform robust and agile bipedal blind locomotion on flat terrain using only their rear legs. This bipedal mode frees the front legs for more complex tasks, significantly enhancing the robot's versatility and functionality. Besides, We evaluate our model on the A1 quadrupedal robot using a biped gait model in the Isaac Gym simulation environment, demonstrating agile and robust movements.

\section{Methodology}
\subsection{Reinforcement Learning on Legged Robots}

The task of learning legged locomotion poses significant challenges due to the complex environment and limitations in sensor data. To address this, a partially observable Markov decision process (POMDP) framework was adopted, denoted as  \((s_t, a_t, P, r_t, p_0, \gamma)\), where \(s_t\) represents the state at time step \(t\) ,\(a_t\) is the action taken by the agent, \(P(s_{t+1}|s_t,a_t)\) describes the system dynamics, predicting the next state based on the current state and action, , \(r_t(s_t,a_t,s_{t+1})\) is the reward function, quantifying the immediate benefit of taking action \(a_t\) in state  \(s_t\) leading to  \(s_{t+1}\), \(p_0\) denotes the initial state distribution, and \(\gamma^t\) is the discount factor, determining the importance of future rewards in the decision-making process.
 The objective of RL in this context is to identify an optimal policy \(\pi_\theta\) parameterized by \(\theta\) , that maximizes the expected discounted return over future trajectories. This is formalized by the objective function \(J(\theta)\):
\begin{equation}
\begin{aligned}
J(\theta) = \mathbb{E}_{\pi_\theta} [\sum^\infty_{t=0}\gamma^{t}r_t]
\label{eq:objective function}
\end{aligned}
\end{equation}

In the state space, the state  \(s_t^{teacher}\) includes proprioceptive observation \(o_t^p \in \mathbb{R}^{48}\), privileged state \(s_t^p \in \mathbb{R}^{45}\) and terrain information \(o^e_t \in \mathbb{R}^{187}\). The proprioceptive observation \(o_t^p \in \mathbb{R}^{48}\) encompasses critical data such as the orientation of the gravity vector, base linear and angular velocity in the robot's frame, joint positions and velocities, the previous action \(a_{t-1} \in \mathbb{R}^{12}\) executed by the current policy. The privileged state \(s_t^p\) contains the information include ground friction coefficients, ground restitution coefficients, contact forces, external forces within positions on the robot, and collision state.
The privileged state \(s_t^p \in \mathbb{R}^{45}\) includes additional key details that the physical robot cannot directly access in the real-world environment. This information encompasses ground friction and restitution coefficients, contact and external forces at specific robot positions, collision state information. The terrain information  \(o^e_t  \in \mathbb{R}^{187}\) contains the 187 height measurement sampled from grid around robot base to the ground.
In contrast, the student policy state utilizes a sequence history of proprioceptive observations \(H_t: [O_{t-N},O_{t-N-1},...,O_{t}]\) to approximate the privileged information. By learning from this historical data, the student policy aims to imitate the inaccessible privileged state, enhancing its decision-making capabilities in the absence of direct access to certain environmental variables.
Regarding the action space, the policy action \(a_t\) is a 12 dimensional target joint position offset added to the time-invariant nominal joint position. This specification guides the joint PD controller, utilizing fixed gains to compute torque commands effectively for motor position control.

\subsection{Adversarial Motion Priors and Rewards Design} \label{AMP}

The AMP framework utilizes adversarial learning to train two neural networks—a generator and a discriminator—in a competitive setup. The generator produces motion predictions for the robot, while the discriminator evaluates their quality and realism. Style rewards are used to measure the similarity between the demonstrator’s behavior and the robot’s, with higher similarity yielding more rewards.

Using the reference dataset \(D\), the AMP-based style reward function encourages the robot to replicate the same gait style. According to \cite{peng2021amp}, a neural network-based discriminator \(D_{\varphi}\) predicts whether a state transition \( (S_{t}, S_{t+1})\) is from the dataset \(D\) or generated by the agent \(A\). Each state \(S_t^{\text{AMP}} \in \mathbb{R}^{31}\) includes joint positions, point velocities, base linear velocity, base angular velocity, and base height relative to the terrain. To avoid mode collapse, the dataset \(D\) contains only trot gait motion clips.

The discriminator's training objective includes a gradient penalty to enforce smoothness and is defined as:
\begin{equation}
\begin{aligned}
\arg\min_{\varphi} & \mathbb{E}_{(s_t,s_{t+1})\sim D}[(D_{\varphi}(s_t,s_{t+1}) - 1)^2] \\
& + \mathbb{E}_{(s_t,s_{t+1})\sim A}[(D_{\varphi}(s_t,s_{t+1}) - 1)^2] \\
& + \frac{\alpha^{qp}}{2} \mathbb{E}_{(s_t,s_{t+1})\sim D}[\|\nabla_{\varphi}D_{\varphi}(s_t,s_{t+1})\|_2^2],
\end{aligned}
\label{eq:Discri}
\end{equation}
where \(\alpha^{qp}\) is a manually specified coefficient (\(\alpha^{qp} = 10\)).

The style reward is defined by:
\begin{equation}
\begin{aligned}
r_t^s[(s_t,s_{t+1}) \sim A] = \mathbb{\max}[0,1-0.25(d^{score}_t - 1)^2],
\end{aligned}
\label{eq:StyleRewardFun}
\end{equation}
where \(d^{score}_t = D_{\varphi}(s_t, s_{t+1})\) and is scaled to the range \([0,1]\).

The overall reward function is:
\begin{equation}
\begin{aligned}
r_t = r_t^g + r_t^s + r_t^l,
\end{aligned}
\label{eq:RewardFun}
\end{equation}
where \(r_t^g\) is the task reward, \(r_t^s\) is the style reward, and \(r_t^l\) is the regularization reward. Task rewards typically include tracking base linear and angular velocities:
\begin{equation}
r_t^g = \omega^v \exp\left(-|\hat{v}_t^{xy} - v_t^{xy}|\right) 
+ \omega^{\omega} \exp\left(-|\hat{\omega}_t^{z} - \omega_t^{z}|\right),
\end{equation}
where \({\omega}^v\) and \({\omega}^\omega\) are coefficients, and \(\hat{v}_t^{xy}\) and \(\hat{\omega}_t^{z}\) are the velocity commands.

Regularization rewards promote safe, smooth motion and minimize energy costs, enhancing the adaptability and efficiency of learned behaviors for real-world applications. This component contributes to the robustness and effectiveness of the overall motion.

\subsection{Reference Generation}

In our research, we refine the imitation and learning of reference data for precise bipedal locomotion. Using a single TO formulation from previous work \cite{winkler18}, we generate walking and running gaits for the A1 biped robot, focusing on its back legs' trajectories.

We streamline this process with TOWR (Trajectory Optimization for Walking Robots) \cite{towr}, which eliminates the need for manual tuning. TOWR generates dynamically feasible, energy-efficient motions by optimizing smooth and stable trajectories. To improve imitation fidelity, we integrate inverse kinematics into TOWR, producing joint space data that closely mimics reference trajectories.

Our generated trajectories encompass various locomotion patterns, including forward walking and two distinct running gaits, each lasting 2.4 seconds. Utilizing TO for motion dataset generation offers several advantages. Firstly, it enables precise matching of the state space between the simulated agent and demonstrator, leveraging kinematic dynamics models to refine trajectory suitability. Moreover, this approach circumvents complexities associated with other motion re-targeting techniques, ensuring a more seamless and accurate replication of desired motions.

\section{Framework and Training}
\subsection{Learning Framework}
\subsubsection{Teacher Policy Architecture}
The teacher policy \(\pi_\theta^{\text{teacher}}\) is trained using Proximal Policy Optimization \cite{schulman2017proximal} with the total reward \(r_t\) as specified in Section \ref{AMP}. During training, the teacher performs a rollout in the environment to generate a state transition \((s_t^{\text{AMP}}, s_{t+1}^{\text{AMP}})\). This state transition is then fed into the discriminator described in Section \ref{AMP}.

The teacher policy consists of three Multilayer Perceptron (MLP) networks. Two of these MLPs encode low-dimensional latent representations: \(l_t^e \in \mathbb{R}^{16}\) for terrain data and \(l_t^p \in \mathbb{R}^{8}\) for privileged data. Using two separate MLPs for encoding helps mitigate information loss that often occurs during the compression process, thereby preserving crucial and necessary information.

The preservation of essential data significantly aids the student policy in reconstructing the latent representations, facilitating a more efficient and accurate learning process. The third MLP acts as a low-level network, utilizing the proprioceptive observation state along with the two encoded latent representations to generate the teacher's action in the environment.
The learning framework is shown in Fig.\ref{fig:Framework}.

\subsubsection{Student Policy Architecture}
The student policy is designed to emulate the teacher policy, replicating actions without relying on privileged state and terrain information. Throughout the student training process, a supervised approach is employed, minimizing two key losses: imitation loss and reconstruction loss. The imitation loss ensures that the student policy closely mimic the action \(a_t^{teacher} \in \mathbb{R}^{12}\) dictated by the teacher's policy. Simultaneously, the reconstruction loss encourages the memory encoder within the student's policy to faithfully reproduce the latent representation \(l_t^{teacher} \in \mathbb{R}^{24}\) consists of the terrain latent  \(l_o^{t} \in \mathbb{R}^{16}\) and privileged latent  \(l_o^{t} \in \mathbb{R}^{8}\) employed by the teacher.

The overarching architecture comprises a memory encoder and a low-level MLP \cite{margolis2022rapid} that maintains an identical structure to the teacher's low-level network. Memory encoders are  implemented by stacking a sequence of 45 historical observations information \(H_t: [O_{t-N},O_{t-N-1},...,O_{t-1}]\) into the input of an MLP.

\subsection{Training and Implementation Details}

\subsubsection{Termination}
The robot's locomotion training is governed by specific termination conditions designed to ensure safety and effectiveness. Episodes are terminated upon detecting collisions involving the trunk, upper limbs, thighs, and calves, prioritizing the study's focus on bipedal locomotion with only two-foot ground contact. 
\subsubsection{Domain Randomization}
To facilitate the transfer of learned behaviors from simulation to the real world, domain randomization has been employed. This approach involves randomizing various parameters crucial for robot locomotion, such as terrain friction, base mass, joint PD controller gains, ground friction, restitution, and perturbations to the robot's base velocity. During training, sampled velocity vectors are added to the robot's current base velocity at random intervals. The specific randomization variables and their corresponding uniform distribution ranges are detailed in Table \ref{Domain Randomization}, enabling robust policy adaptation and testing in diverse real-world environments.

\begin{table}[t]
    \centering
    \small 
    \caption{Randomized Simulation Parameters}
    \label{Domain Randomization}
    \begin{tabular}{|l|l|l|}
        \hline
        \textbf{Parameters} & \textbf{Range} & \textbf{Unit} \\
        \hline
        Link Mass & [0.8, 1.2] & kg \\
        Payload Mass & [0, 3] & kg \\
        Payload Range & [-3] & - \\
        Ground Friction & [0.05, 2.75] & - \\
        Ground Restitution & [0.0, 1.0] & - \\
        Joint $K_p$ & [0.8, 1.2] $\times$ 20 & - \\
        Joint $K_d$ & [0.8, 1.2] $\times$ 0.5 & - \\
        \hline
    \end{tabular}
\end{table}

\begin{figure}
    \centering
    \includegraphics[width=0.8\columnwidth]{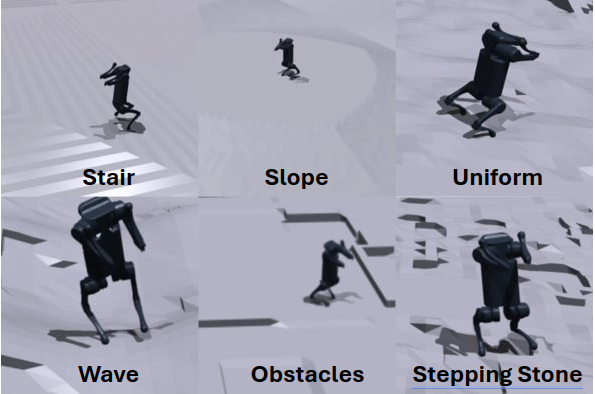}
    \caption{Terrains in Isaac Gym Simulations.}
    \label{fig:Isaac Gym}
    \centering
\end{figure}

\begin{figure*}
    \centering
    \includegraphics[width=0.8\linewidth]{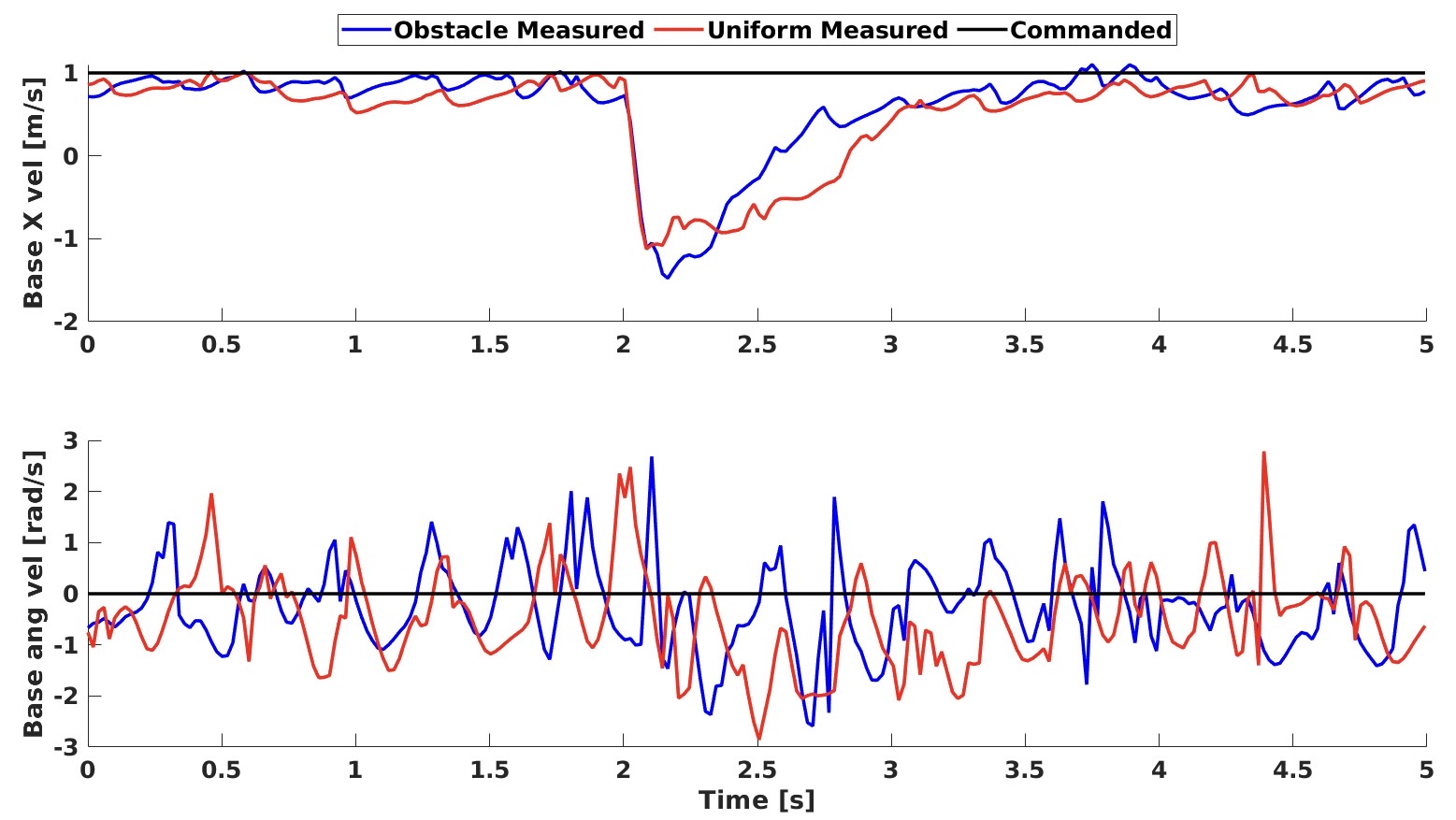}
    \caption{Base linear velocity in the x direction and base angular velocity in yaw for the robot on both uniform and discrete obstacle terrains.}
    \label{fig:base_vel_tracking}
    \centering
\end{figure*}

\subsubsection{Simulation Setup}
We trained 500 parallel agents on different types of terrains with increasing difficulties using the Isaac Gym simulator \cite{rudin2022learning} in overall  26000 iterations and cost 15.88 hours in total. Fig \ref{fig:Isaac Gym} showed the simulation in different terrain type. Each
RL episode lasts for a maximum of 1000 steps,  and terminates early if it reaches the termination criteria. The control frequency of the policy is 50 Hz in the simulation. All training's were performed on a single NVIDIA RTX 4070 GPU.
In A1 locomotion scenarios, actions are represented by  \(a_t \in R^{12}\) a 12-dimensional vector specifying the desired positional adjustments for each actuated joint as dictated by the Proportional-Derivative (PD) controller.

\section{Results}
In our experiments within the Isaac Gym simulation environment, we evaluated the performance of the updated policy for a quadrupedal robot in biped locomotion. The results, depicted in Table \ref{tab:Success Rate}, show varying success rates and tracking accuracy across different terrains and speeds.

\begin{table}
\centering
\caption{Comparison of tracking accuracy and success rate on different terrain and speed}
\label{tab:Success Rate}
\renewcommand{\arraystretch}{1.2} 
\setlength{\tabcolsep}{2pt} 
\scriptsize 
\begin{tabular}{|c|c|>{\centering\arraybackslash}p{1.0cm}|>{\centering\arraybackslash}p{1.0cm}|>{\centering\arraybackslash}p{1.0cm}|>{\centering\arraybackslash}p{1.0cm}|>{\centering\arraybackslash}p{1.0cm}|>{\centering\arraybackslash}p{1.0cm}|}
\hline
\multicolumn{2}{|c|}{\textbf{Terrain Types}} & \textbf{Uniform} & \textbf{Wave} & \textbf{Stepping Stones} & \textbf{Sloped} & \textbf{Stairs} & \textbf{Obstacles} \\
\hline
\multirow{2}{*}{\textbf{0.5 m/s}} & \textbf{Acc (\%)} & 82.76 & 80.84 & 80.06 & 81.37 & 77.67 & 80.88 \\
& \textbf{Succ (\%)} & 92.3 & 91.76 & 91.6 & 84.21 & 90.09 & 89.55 \\
\hline
\multirow{2}{*}{\textbf{1.0 m/s}} & \textbf{Acc (\%)} & 81.29 & 79.64 & 77.73 & 80.01 & 71.57 & 73.51 \\
& \textbf{Succ (\%)} & 90 & 91.59 & 90.01 & 82.14 & 86.14 & 86.59 \\
\hline
\multirow{2}{*}{\textbf{1.5 m/s}} & \textbf{Acc (\%)} & 76.60 & 77.19 & 73.46 & 78.43 & 64.57 & 71.47 \\
& \textbf{Succ (\%)} & 86.96 & 85.06 & 88.73 & 79.5 & 84.91 & 84.55 \\
\hline
\multirow{2}{*}{\textbf{2.0 m/s}} & \textbf{Acc (\%)} & 63.73 & 66.28 & 64.69 & 64.68 & 53.38 & 61.52 \\
& \textbf{Succ (\%)} & 79.55 & 84.22 & 78.55 & 76.23 & 81.53 & 78.1 \\
\hline
\end{tabular}
\end{table}

The robot generally performs well at lower speeds, with higher success rates on less challenging terrains. However, success rates and tracking accuracy decrease as speed and terrain difficulty increase, notably on sloped terrains and obstacles due to mechanical design limitations.

The uniform terrain and discrete obstacle terrain were specifically chosen for detailed analysis as they are highly representative of uneven terrains. The uniform terrain has uneven features, while the discrete obstacle terrain combines elements of both flat and stepped obstacles, akin to stairs, providing a comprehensive challenge for evaluating the robot's locomotion policy.

\begin{figure*}
    \centering
    \includegraphics[width=0.95\linewidth]{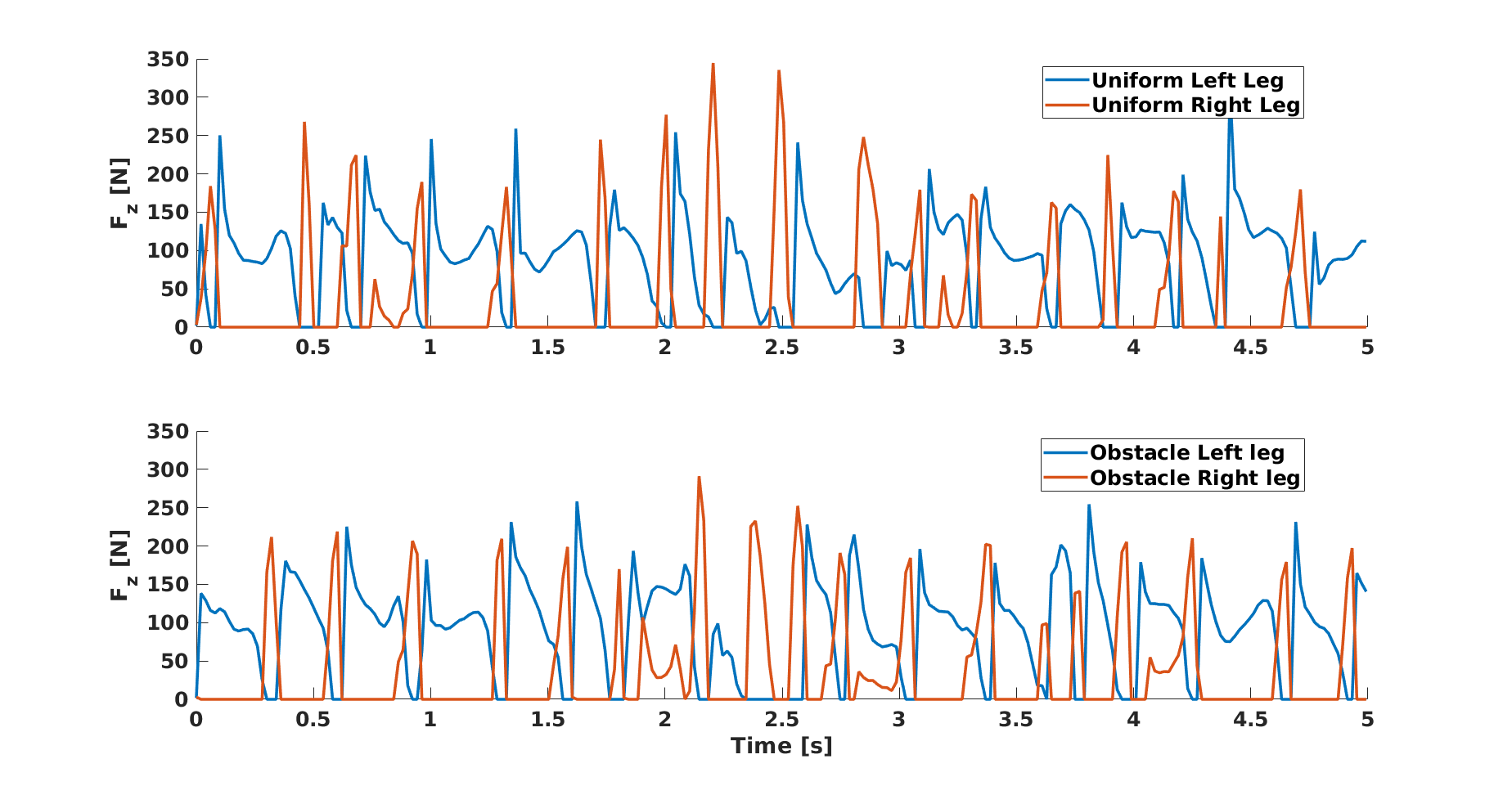}
    \vspace{-2mm}
    \caption{Robot's feet contact forces on both uniform and discrete obstacle terrains.}
    \label{fig:Contact}
    \centering
\end{figure*}

\begin{figure*}
    \centering
    \includegraphics[width=1.0\linewidth]{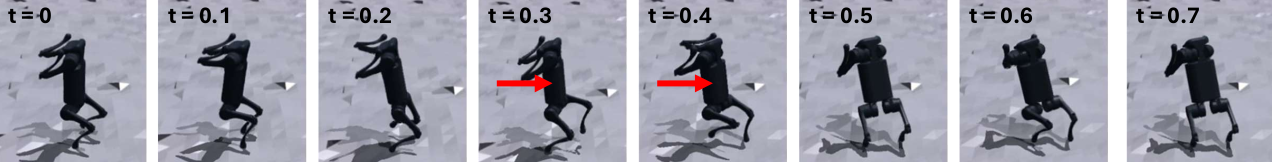}
    \caption{Snapshots illustrating the response of a robot subjected to a 100N push force (indicated by a red arrow) applied along the x-axis over a duration of 0.1 seconds. The sequence shows the robot's movement and stabilization process.}
    \label{fig:push_snapshot}
    \centering
\end{figure*}

Fig. \ref{fig:base_vel_tracking} presents the base linear velocity in the x direction and the base angular velocity in yaw for the robot on both uniform and discrete obstacle terrains. During the push test, noticeable changes in these velocities occur, reflecting the robot's dynamic response to the external force. The robot's base linear velocity tracking shows acceptable performance, with the measured velocities closely following the commanded values. The base angular velocity tracking, while not as accurate as the linear velocity, still demonstrates the robot's ability to follow the commanded trajectory to a reasonable extent. The large spike observed around the 2-second mark is attributed to the external push force. Despite this disturbance, the robot quickly regains stability, indicating robust recovery capabilities of the locomotion policy. This ability to reject disturbances is crucial for maintaining stable locomotion in unpredictable environments.

The contact force data, shown in Fig. \ref{fig:Contact}, provides additional insights into the interaction between the robot's feet and the terrain. The vertical ground reaction forces \(F_z\) were measured for both the rear-left (RL) and rear-right (RR) legs on discrete obstacles and uniform terrains. The data reveals asymmetry in the contact forces, indicating that the learned policy does not exhibit a perfectly symmetrical gait.To address this issue and enhance the robot's performance, future work could incorporate a symmetrical reward function or a reward based on the orientation of base tracking into the learning algorithm.

The push test results and performance analysis provide additional insights into the robot's capabilities and resilience. As illustrated in Fig. \ref{fig:push_snapshot}, the robot's response to a 100N push applied along the x-axis over a duration of 0.1 seconds is captured, showcasing the policy's effectiveness in handling external disturbances. Despite the significant push, the robot manages to stabilize itself quickly, demonstrating the robustness and resilience of the developed locomotion policy.

The results indicate that the legged robot's locomotion policy exhibits satisfactory performance in both base linear velocity tracking and recovery from external disturbances. The ability to maintain stability and follow commanded trajectories on different terrains, as well as the efficient recovery from disturbances, underscores the robustness of the developed policy.

Overall, the experimental outcomes validate the effectiveness of our legged locomotion policy in achieving stable, efficient, and resilient movement across various terrains, even under external disturbances. These findings contribute valuable knowledge towards the development of energy-efficient and robust legged robots capable of operating in diverse and challenging environments.

\section{Conclusion and Future work}

In this paper, we proposed a novel framework for learning robust, agile, and natural bipedal locomotion skills for quadruped robots in simulation. Utilizing a teacher-student learning framework with privileged and terrain information, we enhanced the robustness of the learned policy and helped bridge the sim-to-real gap. By integrating adversarial motion imitation, the learned gait mimics the style and behavior of a TO reference gait. Our results demonstrate high-performance blind locomotion in a quadruped robot in biped mode.

Overall, our findings highlight the potential of imitation learning and TO in achieving agile and robust locomotion across diverse robotic platforms. Future work will focus on developing more robust biped motion capabilities on uneven terrain, transferring these capabilities to physical robots, and refining the transition from quadrupedal to biped mode to enhance legged robots' versatility.

\bibliographystyle{IEEEtran}
\bibliography{Reference}

\begin{thebibliography}{10}
\providecommand{\url}[1]{#1}
\csname url@samestyle\endcsname
\providecommand{\newblock}{\relax}
\providecommand{\bibinfo}[2]{#2}
\providecommand{\BIBentrySTDinterwordspacing}{\spaceskip=0pt\relax}
\providecommand{\BIBentryALTinterwordstretchfactor}{4}
\providecommand{\BIBentryALTinterwordspacing}{\spaceskip=\fontdimen2\font plus
\BIBentryALTinterwordstretchfactor\fontdimen3\font minus \fontdimen4\font\relax}
\providecommand{\BIBforeignlanguage}[2]{{%
\expandafter\ifx\csname l@#1\endcsname\relax
\typeout{** WARNING: IEEEtran.bst: No hyphenation pattern has been}%
\typeout{** loaded for the language `#1'. Using the pattern for}%
\typeout{** the default language instead.}%
\else
\language=\csname l@#1\endcsname
\fi
#2}}
\providecommand{\BIBdecl}{\relax}
\BIBdecl

\bibitem{2024_Bao_Review_bipedal_DRL}
L.~Bao, J.~Humphreys, T.~Peng, and C.~Zhou, ``Deep reinforcement learning for bipedal locomotion: A brief survey,'' 2024.

\bibitem{2016_GAIL_imitation_learning}
J.~Ho and S.~Ermon, ``Generative adversarial imitation learning,'' in \emph{International Conference on Neural Information Processing Systems}, 2016, pp. 4572--4580.

\bibitem{2017_Xuebinpeng_animation_deeploco_DRL_hiarachysystem_referencemotion_deepconvolutionalneuralnewtwork_jointangle_soccer.}
X.~Peng, G.~Berseth, K.~Yin, and M.~Panne, ``{DeepLoco}: {d}ynamic locomotion skills using hierarchical deep reinforcement learning,'' \emph{ACM Transactions on Graphics}, vol.~36, pp. 1--13, 2017.

\bibitem{2020_zhaoming_drl_steppingstones_PPOwithactorcritic_referencefree_simulation}
Z.~Xie, H.~Ling, N.~Kim, and M.~Panne, ``{ALLSTEPS}: Curriculum‐driven learning of stepping stone skills,'' \emph{Computer Graphics Forum}, vol.~39, pp. 213--224, 2020.

\bibitem{2021_siekmann_sim2real_nonreference_perodicreward_DRL_e2e_LSTM_PPO_cassie}
J.~Siekmann, Y.~Godse, A.~Fern, and J.~Hurst, ``Sim-to-real learning of all common bipedal gaits via periodic reward composition,'' in \emph{IEEE International Conference on Robotics and Automation}, 2021, pp. 7309--7315.

\bibitem{2020_Xie_firstsim2real_}
Z.~Xie, P.~Clary, J.~Dao, P.~Morais, J.~Hurst, and M.~van~de Panne, ``Learning locomotion skills for cassie: Iterative design and sim-to-real,'' in \emph{Conference on Robot Learning}, 2020, pp. 317--329.

\bibitem{2021_UCB_hybridrobotics_sim2real_referencebased_HZD_gaitlibrary_e2epolicy_drl_Cassie_lowpassfilter}
Z.~Li, X.~Cheng, X.~B. Peng, P.~Abbeel, S.~Levine, G.~Berseth, and K.~Sreenath, ``Reinforcement learning for robust parameterized locomotion control of bipedal robots,'' in \emph{IEEE International Conference on Robotics and Automation}, 2021, pp. 2811--2817.

\bibitem{2024_Zhang_wholebody_adversarial_motion_priors}
Q.~Zhang, P.~Cui, D.~Yan, J.~Sun, Y.~Duan, A.~Zhang, and R.~Xu, ``Whole-body humanoid robot locomotion with human reference,'' \emph{arXiv preprint arXiv:2402.18294}, 2024.

\bibitem{wu2023learning}
J.~Wu, G.~Xin, C.~Qi, and Y.~Xue, ``Learning robust and agile legged locomotion using adversarial motion priors,'' \emph{IEEE Robotics and Automation Letters}, vol.~8, no.~8, pp. 4975--4982, 2023.

\bibitem{escontrela2022adversarial}
A.~Escontrela, X.~B. Peng, W.~Yu, T.~Zhang, A.~Iscen, K.~Goldberg, and P.~Abbeel, ``Adversarial motion priors make good substitutes for complex reward functions,'' in \emph{IEEE/RSJ International Conference on Intelligent Robots and Systems}, 2022, pp. 25--32.

\bibitem{wang2023learning}
Y.~Wang, Z.~Jiang, and J.~Chen, ``Learning robust, agile, natural legged locomotion skills in the wild,'' in \emph{RoboLetics: Workshop on Robot Learning in Athletics@ CoRL}, 2023.

\bibitem{zhang2024whole}
Q.~Zhang, P.~Cui, D.~Yan, J.~Sun, Y.~Duan, A.~Zhang, and R.~Xu, ``Whole-body humanoid robot locomotion with human reference,'' \emph{arXiv preprint arXiv:2402.18294}, 2024.

\bibitem{winkler18}
A.~Winkler, C.~D. Bellicoso, M.~Hutter, and J.~Buchli, ``Gait and trajectory optimization for legged systems through phase-based end-effector parameterization,'' \emph{IEEE Robotics and Automation Letters}, vol.~3, no.~3, pp. 1560--1567, 2018.

\bibitem{2022_yu_DRL_maneuvers_SRBD_reference_based}
F.~Yu, R.~Batke, J.~Dao, J.~Hurst, K.~Green, and A.~Fern, ``Dynamic bipedal turning through sim-to-real reinforcement learning,'' in \emph{IEEE-RAS International Conference on Humanoid Robots}, 2022, pp. 903--910.

\bibitem{2021_duan_DRL_task-spaceaction_hiarachycotnrolscheme_inversedynamiccontroller}
H.~Duan, J.~Dao, K.~Green, T.~Apgar, A.~Fern, and J.~Hurst, ``Learning task space actions for bipedal locomotion,'' in \emph{IEEE International Conference on Robotics and Automation}, 2021, pp. 1276--1282.

\bibitem{2023_template_taskspace_hierarchyscheme_reducedorderstateALIP_learnedhigherlevel_lowlevelinversedynamiccontroller}
G.~A. Castillo, B.~Weng, S.~Yang, W.~Zhang, and A.~Hereid, ``Template model inspired task space learning for robust bipedal locomotion,'' in \emph{IEEE/RSJ International Conference on Intelligent Robots and Systems}, 2023, pp. 8582--8589.

\bibitem{lee2020learning}
J.~Lee, J.~Hwangbo, L.~Wellhausen, V.~Koltun, and M.~Hutter, ``Learning quadrupedal locomotion over challenging terrain,'' \emph{Science robotics}, vol.~5, no.~47, p. eabc5986, 2020.

\bibitem{kumar2021rma}
A.~Kumar, Z.~Fu, D.~Pathak, and J.~Malik, ``Rma: Rapid motor adaptation for legged robots,'' \emph{arXiv preprint arXiv:2107.04034}, 2021.

\bibitem{2022_multimode_quadrupedtobipedal}
C.~Yu and A.~Rosendo, ``Multi-modal legged locomotion framework with automated residual reinforcement learning,'' \emph{IEEE Robotics and Automation Letters}, vol.~7, no.~4, pp. 10\,312--10\,319, 2022.

\bibitem{peng2021amp}
X.~B. Peng, Z.~Ma, P.~Abbeel, S.~Levine, and A.~Kanazawa, ``{AMP}: Adversarial motion priors for stylized physics-based character control,'' \emph{ACM Transactions on Graphics}, vol.~40, no.~4, pp. 1--20, 2021.

\bibitem{towr}
A.~W. Winkler, ``{TOWR}--an open-source trajectory optimizer for legged robots in c,'' 2018, [Online]. Available: \url{https://github.com/ethz-adrl/towr}.

\bibitem{schulman2017proximal}
J.~Schulman, F.~Wolski, P.~Dhariwal, A.~Radford, and O.~Klimov, ``Proximal policy optimization algorithms,'' \emph{arXiv preprint arXiv:1707.06347}, 2017.

\bibitem{margolis2022rapid}
G.~B. Margolis, G.~Yang, K.~Paigwar, T.~Chen, and P.~Agrawal, ``Rapid locomotion via reinforcement learning,'' \emph{arXiv preprint arXiv:2205.02824}, 2022.

\bibitem{rudin2022learning}
N.~Rudin, D.~Hoeller, P.~Reist, and M.~Hutter, ``Learning to walk in minutes using massively parallel deep reinforcement learning,'' in \emph{Conference on Robot Learning}.\hskip 1em plus 0.5em minus 0.4em\relax PMLR, 2022, pp. 91--100.

\end{thebibliography}

\end{document}